\documentclass{article}
\usepackage{ijcai18}

\usepackage{times}
\usepackage{xcolor}
\usepackage{soul}
\usepackage[utf8]{inputenc}
\usepackage[small]{caption}

\usepackage{helvet}  
\usepackage{courier}  
\usepackage{url}  
\usepackage{graphicx}  
\usepackage{multirow}
\usepackage{amssymb}
\usepackage{amsmath}
\usepackage{subfigure}
\usepackage{dsfont}
\usepackage{array}

\title{Reinforced Mnemonic Reader for Machine Reading Comprehension}

\author{
Minghao Hu$^\dag$\thanks{Contribution during internship at Fudan University and Microsoft Research.}, 
Yuxing Peng$^\dag$, 
Zhen Huang$^\dag$,
Xipeng Qiu$^\ddag$, 
Furu Wei$^\S$, 
Ming Zhou$^\S$
\\ 
$^\dag$ College of Computer, National University of Defense Technology, Changsha, China \\
$^\ddag$ School of Computer Science, Fudan University, Shanghai, China \\
$^\S$ Microsoft Research, Beijing, China  \\
\{huminghao09,pengyuhang,huangzhen\}@nudt.edu.cn \\
xpqiu@fudan.edu.cn,
\{fuwei,mingzhou\}@microsoft.com
}

\begin{document}

\maketitle

\begin{abstract}
 In this paper, we introduce the Reinforced Mnemonic Reader for machine reading comprehension tasks, which enhances previous attentive readers in two aspects.
 First, a reattention mechanism is proposed to refine current attentions by directly accessing to past attentions that are temporally memorized in a multi-round alignment architecture, so as to avoid the problems of \emph{attention redundancy} and \emph{attention deficiency}.
 Second, a new optimization approach, called dynamic-critical reinforcement learning, is introduced to extend the standard supervised method. It always encourages to predict a more acceptable answer so as to address the \emph{convergence suppression} problem occurred in traditional reinforcement learning algorithms. 
 Extensive experiments on the Stanford Question Answering Dataset (SQuAD) show that our model achieves state-of-the-art results. Meanwhile, our model outperforms previous systems by over 6\% in terms of both Exact Match and F1 metrics on two adversarial SQuAD datasets. 
\end{abstract}

\section{Introduction}
\begin{figure}
\begin{center}
\includegraphics[width=3.3in]{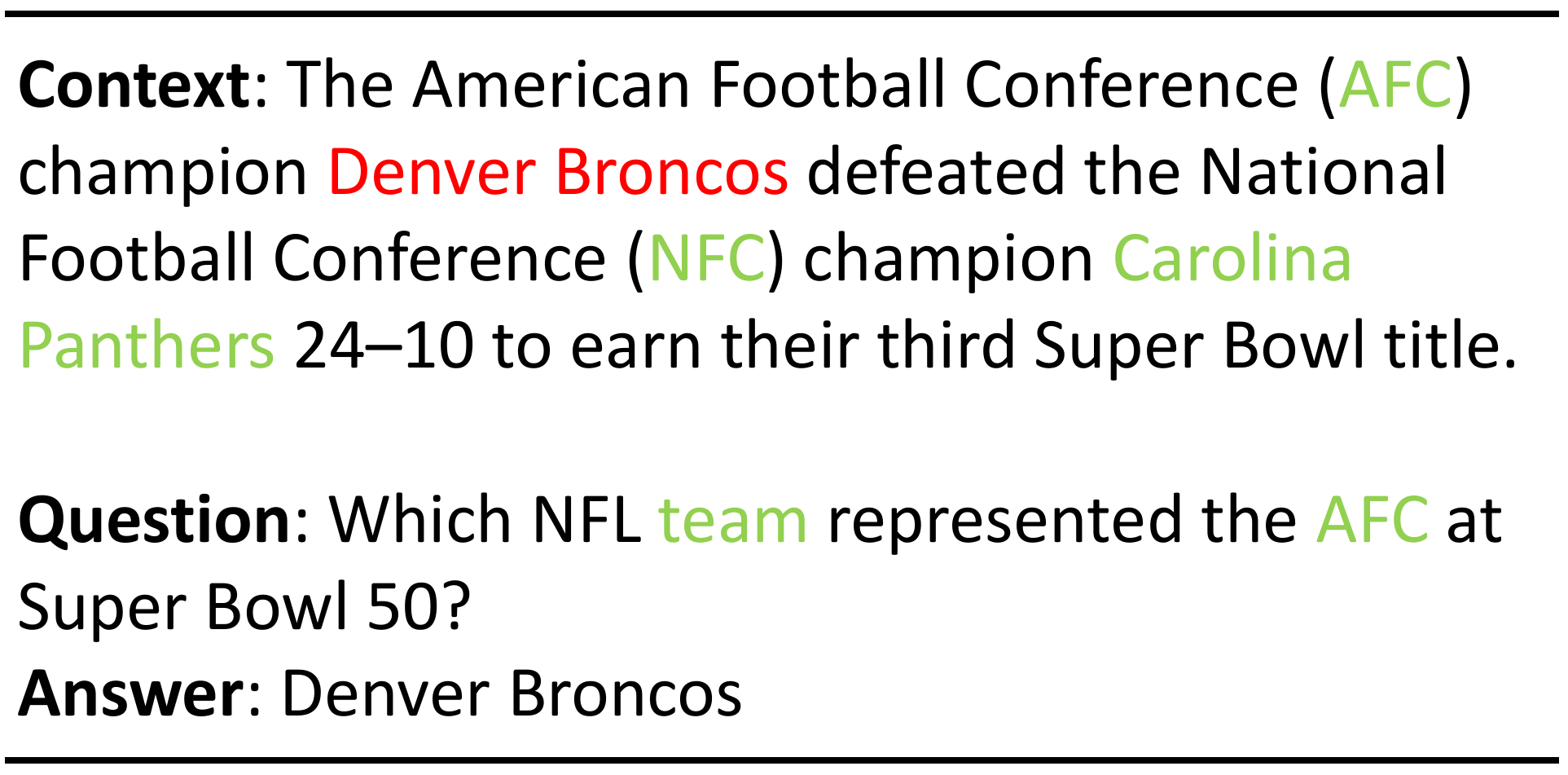}
\end{center}
\caption{An example from the SQuAD dataset. Evidences needed for the answer are marked as green.}
\label{fig4}
\end{figure}

Teaching machines to comprehend a given context paragraph and answer corresponding questions is one of the long-term goals of natural language processing and artificial intelligence. 
Figure \ref{fig4} gives an example of the machine reading comprehension (MRC) task.  
Benefiting from the rapid development of deep learning techniques \cite{goodfellow2016deep} and large-scale benchmark datasets \cite{Hermann15,Hill16,Rajpurkar16}, end-to-end neural networks have achieved promising results on this task \cite{Wang17b,Seo17,Xiong17,Huang17}.

Despite of the advancements, we argue that there still exists two limitations:
\begin{enumerate}
\item To capture complex interactions between the context and the question, a variety of neural attention~\cite{Bahdanau15}, such as bi-attention~\cite{Seo17}, coattention~\cite{Xiong16}, are proposed in a single-round alignment architecture. 
In order to fully compose complete information of the inputs, multi-round alignment architectures that compute attentions repeatedly have been proposed~\cite{Huang17,Xiong17}.
However, in these approaches, the current attention is unaware of which parts of the context and question have been focused in earlier attentions, which results in two distinct but related issues, where multiple attentions 1) focuses on same texts, leading to \emph{attention redundancy} and 2) fail to focus on some salient parts of the input, causing \emph{attention deficiency}.
\item To train the model, standard maximum-likelihood method is used for predicting exactly-matched (EM) answer spans~\cite{Wang17a}.
Recently, reinforcement learning algorithm, which measures the reward as word overlap between the predicted answer and the groung truth, is introduced to optimize towards the F1 metric instead of EM metric~\cite{Xiong17}.
Specifically, an estimated baseline is utilized to normalize the reward and reduce variances. 
However, the convergence can be suppressed when the baseline is better than the reward. This is harmful if the inferior reward is partially overlapped with the ground truth, as the normalized objective will discourage the prediction of ground truth positions.
We refer to this case as the \emph{convergence suppression} problem.
\end{enumerate}

To address the first problem, we present a reattention mechanism that temporally memorizes past attentions and uses them to refine current attentions in a multi-round alignment architecture.
The computation is based on the fact that two words should share similar semantics if their attentions about same texts are highly overlapped, and be less similar vice versa. 
Therefore, the reattention can be more concentrated if past attentions focus on same parts of the input, or be relatively more distracted so as to focus on new regions if past attentions are not overlapped at all.

As for the second problem, we extend the traditional training method with a novel approach called dynamic-critical reinforcement learning. Unlike the traditional reinforcement learning algorithm where the reward and baseline are statically sampled, our approach dynamically decides the reward and the baseline according to two sampling strategies, namely random inference and greedy inference. The result with higher score is always set to be the reward while the other is the baseline. In this way, the normalized reward is ensured to be always positive so that no convergence suppression will be made.

All of the above innovations are integrated into a new end-to-end neural architecture called Reinforced Mnemonic Reader in Figure \ref{fig1}.
We conducted extensive experiments on both the SQuAD~\cite{Rajpurkar16} dataset and two adversarial SQuAD datasets~\cite{Jia17} to evaluate the proposed model. On SQuAD, our single model obtains an exact match (EM) score of 79.5\% and F1 score of 86.6\%, while our ensemble model further boosts the result to 82.3\% and 88.5\% respectively. 
On adversarial SQuAD, our model surpasses existing approahces by more than 6\% on both AddSent and AddOneSent datasets.

\section{MRC with Reattention}
\subsection{Task Description}
For the MRC tasks, a question $Q$ and a context $C$ are given, our goal is to predict an answer $A$, which has different forms according to the specific task. In the SQuAD dataset~\cite{Rajpurkar16}, the answer $A$ is constrained as a segment of text in the context $C$, nerual networks are designed to model the probability distribution $p(A|C, Q)$.

\subsection{Alignment Architecture for MRC}
Among all state-of-the-art works for MRC, one of the key factors is the alignment architecture.
That is, given the hidden representations of question and context, we align each context word with the entire question using attention mechanisms, and enhance the context representation with the attentive question information. 
A detailed comparison of different alignment architectures is shown in Table \ref{table1}.

Early work for MRC, such as Match-LSTM~\cite{Wang17a}, utilizes the attention mechanism stemmed from neural machine translation~\cite{Bahdanau15} serially, where the attention is computed inside the cell of recurrent neural networks.
A more popular approach is to compute attentions in parallel, resulting in a similarity matrix. 
Concretely, given two sets of hidden vectors, $V=\{v_i\}_{i=1}^n$ and $U=\{u_j\}_{j=1}^m$, representing question and context respectively, a similarity matrix $E \in\mathbb{R}^{n \times m}$ is computed as
\begin{eqnarray} \label{eq:1}	
	E_{ij} = f(v_i, u_j) 
\end{eqnarray}
where $E_{ij}$ indicates the similarity between $i$-th question word and $j$-th context word, and $f$ is a scalar function. 
Different methods are proposed to normalize the matrix, resulting in variants of attention such as bi-attention\cite{Seo17} and coattention~\cite{Xiong16}. The attention is then used to attend the question and form a question-aware context representation $H=\{h_j\}_{j=1}^m$.

\begin{table}
\begin{center}
\begin{tabular}{p{2.15cm}<{\centering}p{1cm}<{\centering}p{1cm}<{\centering}p{1.1cm}<{\centering}p{1.1cm}<{\centering}}
\hline 
\multirow{2}*{ \bf Model} & \multicolumn{2}{c}{Aligning Rounds} & Attention \\
 & Interactive & Self &  Type \\ 
\hline
Match-LSTM$^1$ & 1 & - & Serial \\
Rnet$^2$ & 1 & 1 & Serial \\
\hline
BiDAF$^3$ & 1 & - & Parallel \\
FastQAExt$^4$ & 1 & 1 & Parallel \\
DCN+$^5$ & 2 & 2 & Parallel \\
FusionNet$^6$ & 3 & 1 & Parallel \\
\hline
Our Model & 3 & 3 & Parallel \\
\hline
\end{tabular}
\end{center}
\caption{\label{table1} Comparison of alignment architectures of competing models: Wang \& Jiang\protect \shortcite{Wang17a}$^1$, Wang et al.\protect \shortcite{Wang17b}$^2$, Seo et al.\protect \shortcite{Seo17}$^3$, Weissenborn et al.\protect \shortcite{Weissenborn17}$^4$, Xiong et al.\protect \shortcite{Xiong17}$^5$ and Huang et al.\protect \shortcite{Huang17}$^6$.}
\end{table}

Later, Wang et al. \shortcite{Wang17b} propose a serial self aligning method to align the context aginst itself for capturing long-term dependencies among context words. 
Weissenborn et al.~\cite{Weissenborn17} apply the self alignment in a similar way of Eq. \ref{eq:1}, yielding another similarity matrix $B \in\mathbb{R}^{m \times m}$ as
\begin{eqnarray} \label{eq:2}
	B_{ij} = \mathds{1}_{\{i \ne j\}} f(h_i, h_j)  
\end{eqnarray}
where $\mathds{1}_{\{ \cdot \}}$ is an indicator function ensuring that the context word is not aligned with itself. Finally, the attentive information can be integrated to form a self-aware context representation $Z=\{z_j\}_{j=1}^m$, which is used to predict the answer.

We refer to the above process as a single-round alignment architecture. Such architecture, however, is limited in its capability to capture complex interactions among question and context. Therefore, recent works build multi-round alignment architectures by stacking several identical aligning layers~\cite{Huang17,Xiong17}. More specifically, let $V^t=\{v_i^t\}_{i=1}^n$ and $U^t=\{u_j^t\}_{j=1}^m$ denote the hidden representations of question and context in $t$-th layer, and $H^t=\{h_j^t\}_{j=1}^m$ is the corresponding question-aware context representation. Then the two similarity matrices can be computed as
\begin{gather} \label{eq:3}
	E_{ij}^t = f(v_i^t, u_j^t),  \qquad B_{ij}^t = \mathds{1}_{\{i \ne j\}} f(h_i^t, h_j^t)  
\end{gather}

However, one problem is that each alignment is not directly aware of previous alignments in such architecture. 
The attentive information can only flow to the subsequent layer through the hidden representation. 
This can cause two problems: 
1) the \emph{attention redundancy}, where multiple attention distributions are highly similar. Let $\mathrm{softmax}(x)$ denote the softmax function over a vector $x$. Then this problem can be formulized as $D(\mathrm{softmax}(E_{:j}^t) \| \mathrm{softmax}(E_{:j}^k)) < \sigma (t \ne k)$, where $\sigma$ is a small bound and $D$ is a function measuring the distribution distance. 
2) the \emph{attention deficiency}, which means that the attention fails to focus on salient parts of the input: $D(\mathrm{softmax}({E_{:j}^t} ^*) \| \mathrm{softmax}(E_{:j}^t)) > \delta$, where $\delta$ is another bound and $\mathrm{softmax}({E_{:j}^t} ^*)$ is the ``ground truth'' attention distribution.

\subsection{Reattention Mechanism}
\begin{figure}
\begin{center}
\includegraphics[width=3.3in]{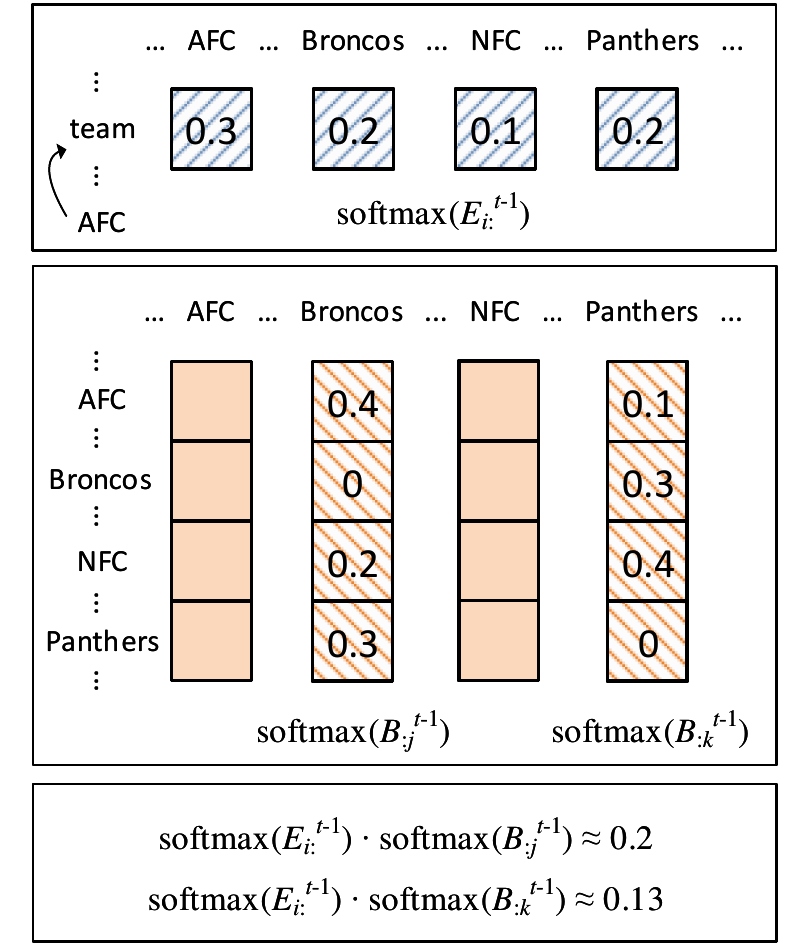}
\end{center}
\caption{Illustrations of reattention for the example in Figure \ref{fig4}.}
\label{fig5}
\end{figure}

To address these problems, we propose to temporally memorize past attentions and explicitly use them to refine current attentions. 
The intuition is that two words should be correlated if their attentions about same texts are highly overlapped, and be less related vice versa.
For example, in Figure \ref{fig5}, suppose that we have access to previous attentions, and then we can compute their dot product to obtain a ``similarity of attention''. In this case, the similarity of word pair (\emph{team}, \emph{Broncos}) is higher than (\emph{team}, \emph{Panthers}).

Therefore, we define the computation of reattention as follows.
Let $E^{t-1}$ and $B^{t-1}$ denote the past similarity matrices that are temporally memorized. The refined similarity matrix $E^t$ ($t > 1$) is computed as
\begin{align} \label{eq:4}
	\tilde{E}_{ij}^t = &\mathrm{softmax}(E_{i:}^{t-1}) \cdot \mathrm{softmax}(B_{:j}^{t-1}) \nonumber \\
	E_{ij}^t =  &f(v_i^t, u_j^t) + \gamma \tilde{E}_{ij}^t
\end{align}
where $\gamma$ is a trainable parameter. Here, $\mathrm{softmax}(E_{i:}^{t-1})$ is the past context attention distribution for the $i$-th question word, and $\mathrm{softmax}(B_{:j}^{t-1})$ is the self attention distribution for the $j$-th context word. 
In the extreme case, when there is no overlap between two distributions, the dot product will be $0$. 
On the other hand, if the two distributions are identical and focus on one single word, it will have a maximum value of $1$. 
Therefore, the similarity of two words can be explicitly measured using their past attentions.
Since the dot product is relatively small than the original similarity, we initialize the $\gamma$ with a tunable hyper-parameter and keep it trainable.
The refined similarity matrix can then be normalized for attending the question.
Similarly, we can compute the refined matrix $B^t$ to get the unnormalized self reattention as
\begin{align} \label{eq:5}
	\tilde{B}_{ij}^t = &\mathrm{softmax}(B_{i:}^{t-1}) \cdot \mathrm{softmax}(B_{:j}^{t-1}) \nonumber \\
	B_{ij}^t = &\mathds{1}_{(i \ne j)} \left( f(h_i^t, h_j^t) + \gamma \tilde{B}_{ij}^t \right) 
\end{align}

\section{Dynamic-critical Reinforcement Learning}
In the extractive MRC task, the model distribution $p(A|C, Q;\theta)$ can be divided into two steps: first predicting the start position $i$ and then the end position $j$ as
\begin{eqnarray}	\label{eq:6}
	p(A|C,Q;\theta) = p_1(i|C,Q;\theta) p_2(j|i,C,Q;\theta)
\end{eqnarray}
where $\theta$ represents all trainable parameters.

The standard maximum-likelihood (ML) training method is to maximize the log probabilities of the ground truth answer positions~\cite{Wang17a}
\begin{eqnarray} \label{eq:7}
	\mathcal{L}_{ML}(\theta) = -\sum_{k} \log p_1(y_k^1) + \log p_2(y_k^2|y_k^1)
\end{eqnarray}
where $y_k^1$ and $y_k^2$ are the answer span for the $k$-th example, and we denote $p_1(i|C,Q;\theta)$ and $p_2(j|i,C,Q;\theta)$ as $p_1(i)$ and $p_2(j|i)$ respectively for abbreviation.

Recently, reinforcement learning (RL), with the task reward measured as word overlap between predicted answer and groung truth, is introduced to MRC~\cite{Xiong17}.
A baseline $b$, which is obtained by running greedy inference with the current model, is used to normalize the reward and reduce variances. Such approach is known as the self-critical sequence training (SCST)~\cite{Rennie2016}, which is first used in image caption. 
More specifically, let $R(A^s, A^*)$ denote the F1 score between a sampled answer $A^s$ and the ground truth $A^*$. 
The training objective is to minimize the negative expected reward by
\begin{eqnarray} \label{eq:8}
	\mathcal{L}_{SCST}(\theta) = -\mathbb{E}_{A^s \sim p_{\theta}(A)}[R(A^s)-R(\hat{A})]
\end{eqnarray}
where we abbreviate the model distribution $p(A|C,Q;\theta)$ as $p_{\theta}(A)$, and the reward function $R(A^s, A^*)$ as $R(A^s)$. 
$\hat{A}$ is obtained by greedily maximizing the model distribution: \begin{displaymath}\hat{A} = \arg\max_{A}p(A|C,Q;\theta)\end{displaymath}

The expected gradient $\nabla_{\theta}\mathcal{L}_{SCST}(\theta)$ can be computed according to the REINFORCE algorithm~\cite{Sutton98} as
\begin{align} \label{eq:9}
	\nabla_{\theta}\mathcal{L}_{SCST}(\theta) &= -\mathbb{E}_{A^s \sim p_{\theta}(A)}[ \left( R(A^s)-b \right) \nabla_{\theta}\log p_{\theta}(A^s)] \nonumber \\
	&\approx - \left( R(A^s)-R(\hat{A}) \right) \nabla_{\theta}\log p_{\theta}(A^s)
\end{align}
where the gradient can be approxiamated using a single Monte-Carlo sample $A^s$ derived from $p_{\theta}$.

However, a sampled answer is discouraged by the objective when it is worse than the baseline. 
This is harmful if the answer is partially overlapped with ground truth, since the normalized objective would discourage the prediction of ground truth positions.
For example, in Figure \ref{fig4}, suppose that $A^s$ is \emph{champion Denver Broncos} and $\hat{A}$ is \emph{Denver Broncos}. Although the former is an acceptable answer, the normalized reward would be negative and the prediction for end position would be suppressed, thus hindering the convergence. 
We refer to this case as the \emph{convergence suppression} problem.

Here, we consider both random inference and greedy inference as two different sampling strategies: the first one encourages exploration while the latter one is for exploitation\footnote{In practice we found that a better approximation can be made by considering a top-$K$ answer list, where $\hat{A}$ is the best result and $A^s$ is sampled from the rest of the list.}. Therefore, we approximate the expected gradient by dynamically set the reward and baseline based on the F1 scores of both $A^s$ and $\hat{A}$. The one with higher score is set as reward, while the other is baseline. We call this approach as dynamic-critical reinforcement learning (DCRL)
\begin{align} \label{eq:10}
	\nabla&_{\theta}\mathcal{L}_{DCRL}(\theta) = -\mathbb{E}_{A^s \sim p_{\theta}(A)}[ \left( R(A^s)-b \right) \nabla_{\theta}\log p_{\theta}(A^s)] \nonumber \\
	& \approx -\mathds{1}_{\{R(A^s) \ge R(\hat{A})\}} \left( R(A^s)-R(\hat{A}) \right) \nabla_{\theta}\log p_{\theta}(A^s) \nonumber \\
	& -\mathds{1}_{\{R(\hat{A}) > R(A^s)\}} \left( R(\hat{A})-R(A^s) \right) \nabla_{\theta}\log p_{\theta}(\hat{A})
\end{align}

Notice that the normalized reward is constantly positive so that superior answers are always encouraged. Besides, when the score of random inference is higher than the greedy one, DCRL is equivalent to SCST. Thus, Eq. \ref{eq:9} is a special case of Eq. \ref{eq:10}.

Following~\cite{Xiong17} and~\cite{Kendall17}, we combine ML and DCRL objectives using homoscedastic uncertainty as task-dependent weightings so as to stabilize the RL training as
\begin{eqnarray}	\label{eq:11}
	\mathcal{L} = \frac{1}{2 \sigma_a^2} \mathcal{L}_{ML} + \frac{1}{2 \sigma_b^2} \mathcal{L}_{DCRL} + \log \sigma_a^2 + \log \sigma_b^2
\end{eqnarray}
where $\sigma_a$ and $\sigma_b$ are trainable parameters.

\section{End-to-end Architecture}
\begin{figure}
\begin{center}
\includegraphics[width=3.3in]{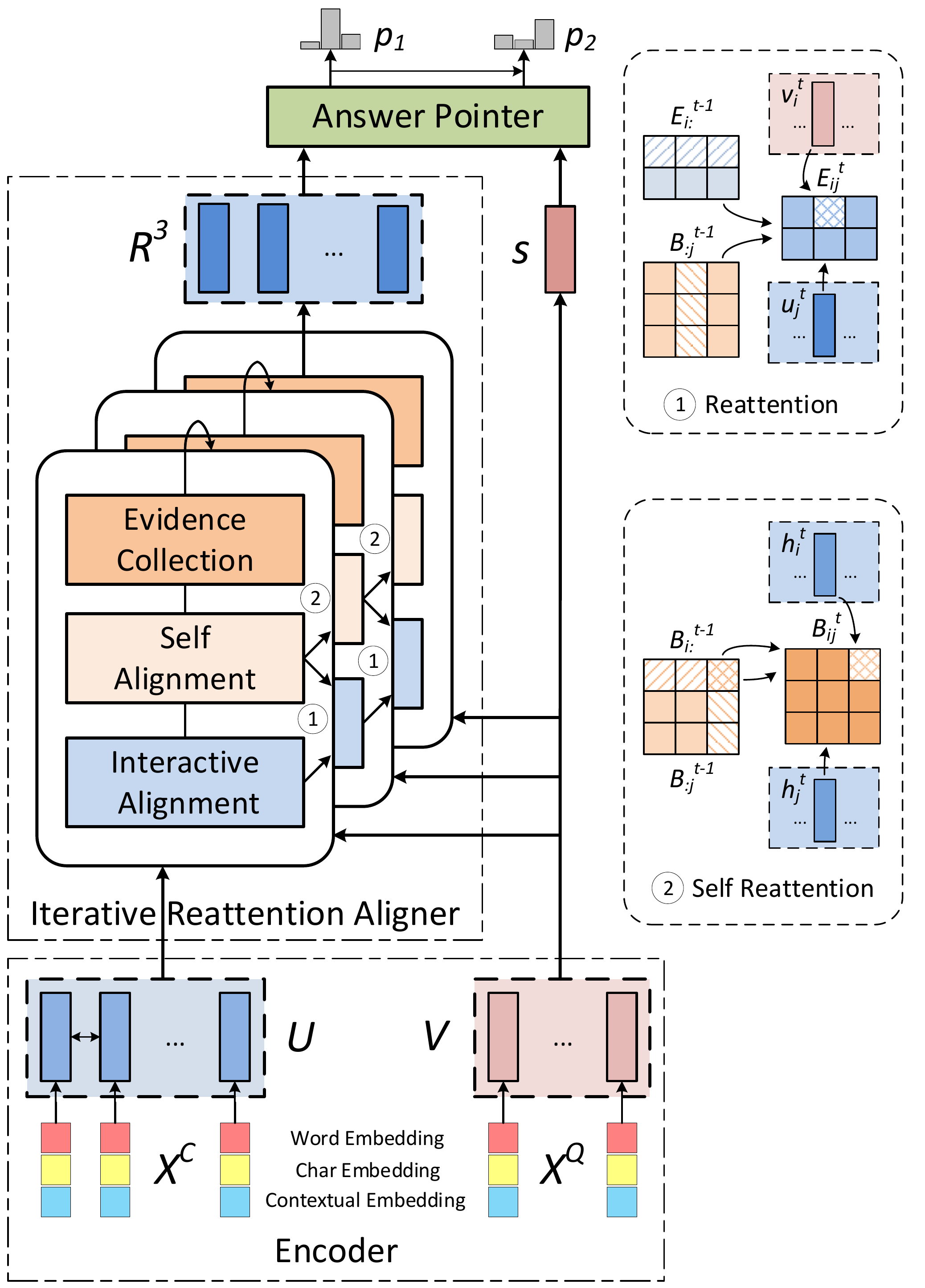}
\end{center}
\caption{The architecture overview of Reinforced Mnemonic Reader. The subfigures to the right show detailed demonstrations of the reattention mechanism: 1) refined $E^t$ to attend the query; 2) refined $B^t$ to attend the context.}
\label{fig1}
\end{figure}

Based on previous innovations, we introduce an end-to-end architecture called Reinforced Mnemonic Reader, which is shown in Figure \ref{fig1}. 
It consists of three main components: 
1) an encoder builds contextual representations for question and context jointly; 
2) an iterative aligner performs multi-round alignments between question and context with the reattention mechanism; 
3) an answer pointer predicts the answer span sequentially.
Beblow we give more details of each component.

\noindent{\bf Encoder.} 
Let $W^Q=\{w_i^q\}_{i=1}^n$ and $W^C=\{w_j^c\}_{j=1}^m$ denote the word sequences of the question and context respectively. 
The encoder firstly converts each word to an input vector.
We utilize the 100-dim GloVe embedding~\cite{Pennington14} and 1024-dim ELMo embedding~\cite{Elmo17}. Besides, a character-level embedding is obtained by encoding the character sequence with a bi-directional long short-term memory network (BiLSTM)~\cite{Hochreiter97}, where two last hidden states are concatenated to form the embedding.
In addition, we use binary feature of exact match, POS embedding and NER embedding for both question and context, as suggested in~\cite{Chen17a}. 
Together the inputs $X^Q=\{x_i^q\}_{i=1}^n$ and $X^C=\{x_j^c\}_{j=1}^m$ are obtained.

To model each word with its contextual information, a weight-shared BiLSTM is utilized to perform the encoding
\begin{gather} \label{eq:12}
	v_i=\mathrm{BiLSTM}(x_i^q) , \quad  u_j=\mathrm{BiLSTM}(x_j^c)
\end{gather}
Thus, the contextual representations for both question and context words can be obtained, denoted as two matrices: $V=[v_1,...,v_n] \in\mathbb{R}^{2d \times n}$ and $U=[u_1,...,u_m] \in\mathbb{R}^{2d \times m}$.

\begin{figure}
\begin{center}
\includegraphics[width=3.3in]{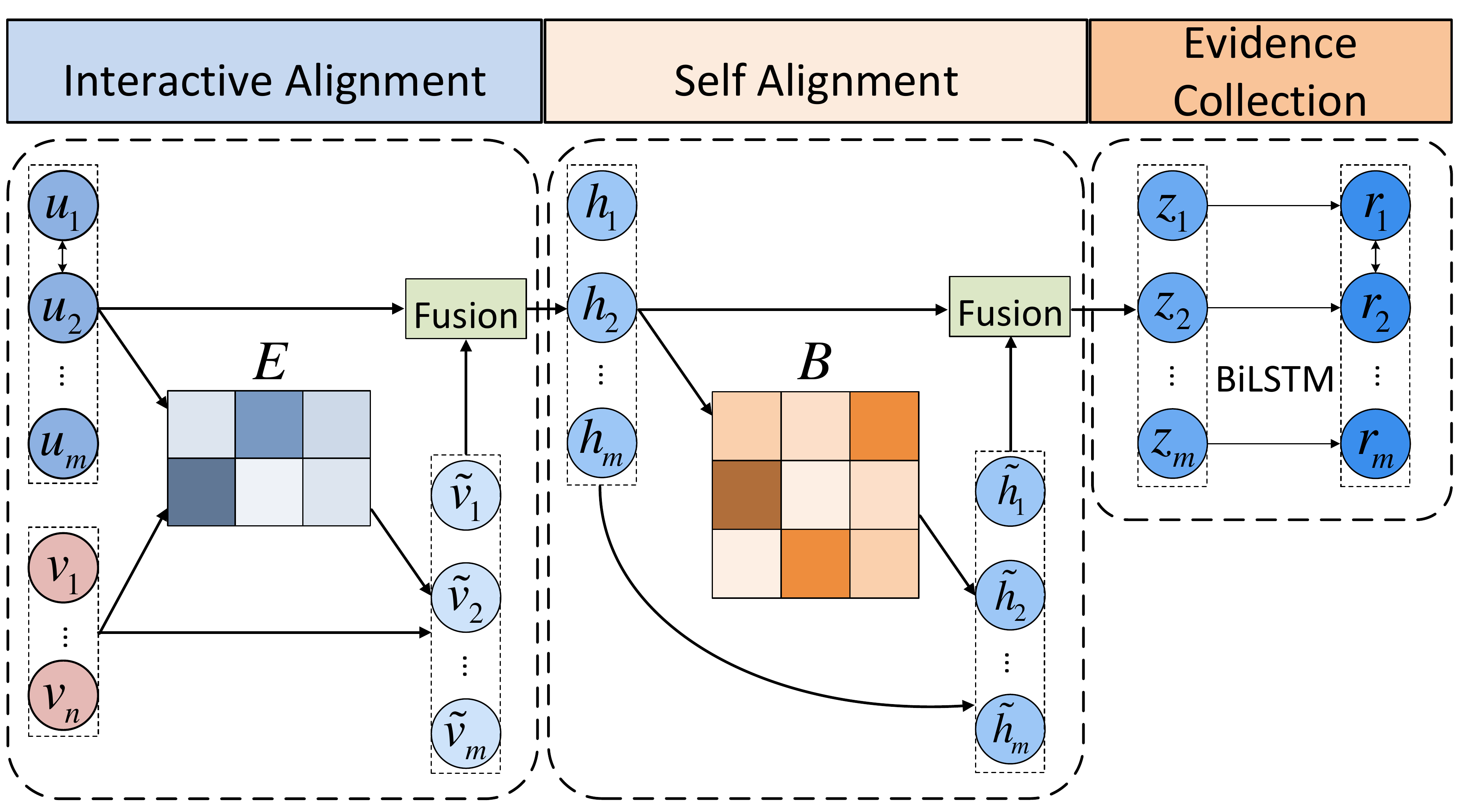}
\end{center}
\caption{The detailed overview of a single aligning block. Different colors in $E$ and $B$ represent different degrees of similarity.}
\label{fig2}
\end{figure}

\noindent{\bf Iterative Aligner.} 
The iterative aligner contains a stack of three aligning blocks. 
Each block consists of three modules:
1) an interactive alignment to attend the question into the context;
2) a self alignment to attend the context against itself;
3) an evidence collection to model the context representation with a BiLSTM.
The reattention mechanism is utilized between two blocks, where past attentions are temporally memorizes to help modulating current attentions.
Below we first describe a single block in details, which is shown in Figure \ref{fig2}, and then introduce the entire architecture.

\noindent{\bf Single Aligning Block.} 
First, the similarity matrix $E \in\mathbb{R}^{n \times m}$ is computed using Eq. \ref{eq:1}, where the multiplicative product with nonlinearity is applied as attention function: $f(u,v) = \mathrm{relu}(W_u u) ^ \mathsf{T} \mathrm{relu}(W_v v)$.
The question attention for the $j$-th context word is then: $\mathrm{softmax}(E_{:j})$, which is used to compute an attended question vector $\tilde{v}_j = V \cdot \mathrm{softmax}(E_{:j})$.

To efficiently fuse the attentive information into the context, an heuristic fusion function, denoted as $o=\mathrm{fusion}(x, y)$, is proposed as
\begin{gather} \label{eq:13}
\tilde{x} = \mathrm{relu} \left(W_r [x; y; x \circ y; x - y] \right) \nonumber \\
g = \sigma \left(W_g [x; y; x \circ y; x - y] \right) \nonumber \\
o = g \circ \tilde{x}  + (1 - g) \circ x 
\end{gather}
where $\sigma$ denotes the sigmoid activation function, $\circ$ denotes element-wise multiplication, and the bias term is omitted. 
The computation is similar to the highway networks~\cite{Srivastava15}, where the output vector $o$ is a linear interpolation of the input $x$ and the intermediate vector $\tilde{x}$. A gate $g$ is used to control the composition degree to which the intermediate vector is exposed.
With this function, the question-aware context vectors $H=[h_1,...,h_m]$ can be obtained as: $h_j = \mathrm{fusion}(u_j, \tilde{v}_j)$.

Similar to the above computation, a self alignment is applied to capture the long-term dependencies among context words. Again, we compute a similarity matrix $B \in\mathbb{R}^{m \times m}$ using Eq. \ref{eq:2}. The attended context vector is then computed as: $\tilde{h}_j = H \cdot \mathrm{softmax}(B_{:j})$, where $\mathrm{softmax}(B_{:j})$ is the self attention for the $j$-th context word. Using the same fusion function as $z_j = \mathrm{fusion}(h_j, \tilde{h}_j)$, we can obtain self-aware context vectors $Z=[z_1,...,z_m]$.

Finally, a BiLSTM is used to perform the evidence collection, which outputs the fully-aware context vectors $R=[r_1,...,r_m]$ with $Z$ as its inputs.

\noindent{\bf Multi-round Alignments with Reattention.} 
To enhance the ability of capturing complex interactions among inputs, we stack two more aligning blocks with the reattention mechanism as follows

\begin{gather} \label{eq:14}
	R^1, Z^1, E^1, B^1 = \mathrm{align}^1(U, V)	\nonumber \\
	R^2, Z^2, E^2, B^2 = \mathrm{align}^2(R^1, V, E^1, B^1)	\nonumber \\
	R^3 ,Z^3, E^3, B^3 = \mathrm{align}^3(R^2, V, E^2, B^2, Z^1, Z^2)
\end{gather}
where $\mathrm{align}^t$ denote the $t$-th block. 
In the $t$-th block ($t>1$), we fix the hidden representation of question as $V$, and set the hidden representation of context as previous fully-aware context vectors $R^{t-1}$. Then we compute the unnormalized reattention $E^t$ and $B^t$ with Eq. \ref{eq:4} and Eq. \ref{eq:5} respectively.
In addition, we utilize a residual connection~\cite{He16} in the last BiLSTM to form the final fully-aware context vectors $R^3=[r_1^3,...,r_m^3]$: $r_j^3=\mathrm{BiLSTM} \left([z_j^1; z_j^2; z_j^3] \right)$.

\noindent{\bf Answer Pointer.}
We apply a variant of pointer networks~\cite{Vinyals15} as the answer pointer to make the predictions.
First, the question representation $V$ is summarized into a fixed-size summary vector $s$ as: $s=\sum_{i=1}^n \alpha_i v_i$, where $\alpha_i \propto \mathrm{exp}(w^ \mathrm{ T } v_i)$.
Then we compute the start probability $p_1(i)$ by heuristically attending the context representation $R^3$ with the question summary $s$ as
\begin{eqnarray} \label{eq:15}
	p_1(i) \propto \mathrm{exp} \left( w_1^\mathrm{T} \mathrm{tanh}(W_1[r_i^3; s; r_i^3 \circ s; r_i^3 - s]) \right)
\end{eqnarray}

Next, a new question summary $\tilde{s}$ is updated by fusing context information of the start position, which is computed as $l=R^3 \cdot p_1$, into the old question summary: $\tilde{s} = \mathrm{fusion}(s, l)$. 
Finally the end probability $p_2(j|i)$ is computed as
\begin{eqnarray} \label{eq:16}
	p_2(j|i) \propto \mathrm{exp} \left( w_2^\mathrm{T} \mathrm{tanh}(W_2[r_j^3; \tilde{s}; r_j^3 \circ \tilde{s}; r_j^3 - \tilde{s}]) \right)
\end{eqnarray}

\section{Experiments}
\subsection{Implementation Details}
We mainly focus on the SQuAD dataset~\cite{Rajpurkar16} to train and evaluate our model. SQuAD is a machine comprehension dataset, totally containing more than $100,000$ questions manually annotated by crowdsourcing workers on a set of $536$ Wikipedia articles. In addition, we also test our model on two adversarial SQuAD datasets~\cite{Jia17}, namely AddSent and AddOneSent. In both adversarial datasets, a confusing sentence with a wrong answer is appended at the end of the context in order to fool the model.

We evaluate the Reinforced Mnemonic Reader (R.M-Reader) by running the following setting. We first train the model until convergence by optimizing Eq. \ref{eq:7}. We then finetune this model with Eq. \ref{eq:11}, until the F1 score on the development set no longer improves.

We use the Adam optimizer~\cite{Kingma14} for both ML and DCRL training. The initial learning rates are $0.0008$ and $0.0001$ respectively, and are halved whenever meeting a bad iteration. The batch size is $48$ and a dropout rate~\cite{Srivastava14} of $0.3$ is used to prevent overfitting. Word embeddings remain fixed during training. For out of vocabulary words, we set the embeddings from Gaussian distributions and keep them trainable. The size of character embedding and corresponding LSTMs is $50$, the main hidden size is $100$, and the hyperparameter $\gamma$ is $3$.

\subsection{Overall Results}
We submitted our model on the hidden test set of SQuAD for evaluation. 
Two evaluation metrics are used: Exact Match (EM), which measures whether the predicted answer are exactly matched with the ground truth, and F1 score, which measures the degree of word overlap at token level. 

As shown in Table \ref{table3}, R.M-Reader achieves an EM score of 79.5\% and F1 score of 86.6\%. Since SQuAD is a competitve MRC benchmark, we also build an ensemble model that consists of $12$ single models with the same architecture but initialized with different parameters. Our ensemble model improves the metrics to 82.3\% and 88.5\% respectively\footnote{The results are on https://worksheets.codalab.org/worksheets/ 0xe6c23cbae5e440b8942f86641f49fd80.}.

\begin{table}
\begin{center}
\begin{tabular}{lllll}
\hline 
\multirow{2}*{ \emph{Single Model} } & \multicolumn{2}{c}{ \textbf{Dev} } & \multicolumn{2}{c}{ \textbf{Test} } \\
 & \bf EM & \bf F1 & \bf EM & \bf F1 \\ 
\hline
LR Baseline$^1$ & 40.0 & 51.0 & 40.4 & 51.0 \\ 
DCN+$^2$ & 74.5 & 83.1 & 75.1 & 83.1 \\
FusionNet$^3$ & 75.3 & 83.6 & 76.0 & 83.9 \\
SAN$^4$ & 76.2 & 84.1 & 76.8 & 84.4 \\
AttentionReader+$^\dagger$ & - & - & 77.3 & 84.9 \\
BSE$^5$ & 77.9 & 85.6 & 78.6 & 85.8 \\
R-net+$^\dagger$ & - & - & 79.9 & 86.5 \\
SLQA+$^\dagger$ & - & - & 80.4 & 87.0 \\
Hybrid AoA Reader+$^\dagger$ & - & - & 80.0 & 87.3 \\
\bf R.M-Reader & \bf 78.9 & \bf 86.3 & \bf 79.5 & \bf 86.6 \\
\hline
\emph{Ensemble Model} &  & \\ 
DCN+$^2$ & - & - & 78.8 & 86.0 \\
FusionNet$^3$ & 78.5 & 85.8 & 79.0 & 86.0 \\
SAN$^4$ & 78.6 & 85.8 & 79.6 & 86.5 \\
BSE$^5$ & 79.6 & 86.6 & 81.0 & 87.4 \\
AttentionReader+$^\dagger$ & - & - & 81.8 & 88.2 \\
R-net+$^\dagger$ & - & - & 82.6 & 88.5 \\
SLQA+$^\dagger$ & - & - & 82.4 & 88.6 \\
Hybrid AoA Reader+$^\dagger$ & - & - & 82.5 & 89.3 \\
\bf R.M-Reader & \bf 81.2 & \bf 87.9 & \bf 82.3 & \bf 88.5 \\
\hline
Human$^1$ & 80.3 & 90.5 & 82.3 & 91.2 \\
\end{tabular}
\end{center}
\caption{\label{table3} The performance of Reinforced Mnemonic Reader and other competing approaches on the SQuAD dataset. The results of test set are extracted on Feb 2, 2018: Rajpurkar et al.\protect \shortcite{Rajpurkar16}$^1$, Xiong et al.\protect \shortcite{Xiong17}$^2$, Huang et al.\protect \shortcite{Huang17}$^3$, Liu et al.\protect \shortcite{Liu17}$^4$ and Peters\protect \shortcite{Elmo17}$^5$. $\dagger$ indicates unpublished works. BSE refers to BiDAF + Self Attention + ELMo.}
\end{table}

Table \ref{table4} shows the performance comparison on two adversarial datasets, AddSent and AddOneSent. All models are trained on the original train set of SQuAD, and are tested on the two datasets. As we can see, R.M-Reader comfortably outperforms all previous models by more than 6\% in both EM and F1 scores, indicating that our model is more robust against adversarial attacks.

\subsection{Ablation Study}
The contributions of each component of our model are shown in Table \ref{table5}. 
Firstly, ablation (1-4) explores the utility of reattention mechanism and DCRL training method.
We notice that reattention has more influences on EM score while DCRL contributes more to F1 metric, and removing both of them results in huge drops on both metrics. 
Replacing DCRL with SCST also causes a marginal decline of performance on both metrics.
Next, we relace the default attention function with the dot product: $f(u,v)=u \cdot v$ (5), and both metrics suffer from degradations.
(6-7) shows the effectiveness of heuristics used in the fusion function. Removing any of the two heuristics leads to some performance declines, and heuristic subtraction is more effective than multiplication.
Ablation (8-9) further explores different forms of fusion, where gate refers to $o=g \circ \tilde{x}$ and MLP denotes $o=\tilde{x}$ in Eq. \ref{eq:13}, respectively.
In both cases the highway-like function has outperformed its simpler variants.
Finally, we study the effect of different numbers of aligning blocks in (10-12). We notice that using 2 blocks causes a slight performance drop, while increasing to 4 blocks barely affects the SoTA result. Interestingly, a very deep alignment with 5 blocks results in a significant performance decline. We argue that this is because the model encounters the degradation problem existed in deep networks~\cite{He16}.

\begin{table}
\begin{center}
\begin{tabular}{lllll}
\hline 
\multirow{2}*{ \emph{Model} } & \multicolumn{2}{c}{ \textbf{AddSent} } & \multicolumn{2}{c}{ \textbf{AddOneSent} } \\
 & \bf EM & \bf F1 & \bf EM & \bf F1 \\ 
\hline
LR Baseline & 17.0 & 23.2 & 22.3 & 41.8 \\ 
Match-LSTM$^{1\ast}$ & 24.3 & 34.2 & 34.8 & 41.8 \\ 
BiDAF$^{2\ast}$ & 29.6 & 34.2 & 40.7 & 46.9 \\
SEDT$^{3\ast}$ & 30.0 & 35.0 & 40.0 & 46.5 \\
ReasoNet$^{4\ast}$ & 34.6 & 39.4 & 43.6 & 49.8 \\ 
FusionNet$^{5\ast}$ & 46.2 & 51.4 & 54.7 & 60.7 \\
\bf R.M-Reader & \bf 53.0 & \bf 58.5 & \bf 60.9 & \bf 67.0 \\
\hline 
\end{tabular}
\end{center}
\caption{\label{table4} Performance comparison on two adversarial SQuAD datasets. Wang \& Jiang\protect \shortcite{Wang17a}$^1$, Seo et al.\protect \shortcite{Seo17}$^2$, Liu et al.\protect \shortcite{Liu17b}$^3$, Shen et al.\protect \shortcite{Shen16}$^4$ and Huang et al.\protect \shortcite{Huang17}$^5$. $\ast$ indicates ensemble models.}
\end{table}

\begin{table}
\begin{center}
\begin{tabular}{llllll}
\hline 
\textbf{Configuration} & \bf EM & \bf F1 & \bf $\Delta$EM & \bf $\Delta$F1 \\ 
\hline
R.M-Reader & 78.9 & 86.3 & $-$ & $-$ \\
\hline
(1) - Reattention & 78.1 & 85.8 & -0.8 & -0.5 \\
(2) - DCRL & 78.2 & 85.4 & -0.7 & -0.9 \\
(3) - Reattention, DCRL & 77.1 & 84.8 & -1.8 & -1.5 \\
(4) - DCRL, + SCST & 78.5 & 85.8 & -0.4 & -0.5 \\
(5) Attention: Dot & 78.2 & 85.9 & -0.7 & -0.4 \\
(6) - Heuristic Sub & 78.1 & 85.7 & -0.8 & -0.6 \\
(7) - Heuristic Mul & 78.3 & 86.0 & -0.6 & -0.3 \\
(8) Fusion: Gate  & 77.9 & 85.6 & -1.0 & -0.7 \\
(9) Fusion: MLP  & 77.2 & 85.2 & -1.7 & -1.1 \\
(10) Num of Blocks: 2 & 78.7 & 86.1 & -0.2 & -0.2 \\
(11) Num of Blocks: 4 & 78.8 & 86.3 & -0.1 & 0 \\
(12) Num of Blocks: 5 & 77.5 & 85.2 & -1.4 & -1.1 \\
\hline
\end{tabular}
\end{center}
\caption{\label{table5} Ablation study on SQuAD dev set.}
\end{table}

\subsection{Effectiveness of Reattention}
We further present experiments to demonstrate the effectiveness of reattention mechanism. For the attention redundancy problem, we measure the distance of attention distributions in two adjacent aligning blocks, e.g., $\mathrm{softmax}(E_{:j}^1)$ and $\mathrm{softmax}(E_{:j}^2)$. Higher distance means less attention redundancy. For the attention deficiency problem, we take the arithmetic mean of multiple attention distributions from the ensemble model as the ``ground truth'' attention distribution $\mathrm{softmax}({E_{:j}^t} ^*)$, and compute the distance of individual attention $\mathrm{softmax}(E_{:j}^t)$ with it. Lower distance refers to less attention deficiency. We use Kullback–Leibler divergence as the distance function $D$, and we report the averaged value over all examples.

\begin{table}
\begin{center}
\begin{tabular}{lcc}
\hline 
\textbf{KL divergence} & \bf - Reattention & \bf + Reattention \\ 
\hline
\emph{Redundancy} &  & \\ 
$E^1$ to $E^2$ & 0.695 $\pm$ 0.086 & \textbf{0.866} $\pm$ 0.074 \\
$E^2$ to $E^3$ & 0.404 $\pm$ 0.067 & 0.450 $\pm$ 0.052 \\
$B^1$ to $B^2$ & 0.976 $\pm$ 0.092 & \textbf{1.207} $\pm$ 0.121 \\
$B^2$ to $B^3$ & 1.179 $\pm$ 0.118 & 1.193 $\pm$ 0.097 \\
\hline
\emph{Deficiency} &  & \\ 
$E^2$ to ${E^2} ^*$ & 0.650 $\pm$ 0.044 & \textbf{0.568} $\pm$ 0.059 \\
$E^3$ to ${E^3} ^*$ & 0.536 $\pm$ 0.047 & \textbf{0.482} $\pm$ 0.035\\
\hline
\end{tabular}
\end{center}
\caption{\label{table7} Comparison of KL diverfence on different attention distributions on SQuAD dev set.}
\end{table}

Table \ref{table7} shows the results. We first see that the reattention indeed help in alleviating the attention redundancy: the divergence between any two adjacent blocks has been successfully enlarged with reattention. However, we find that the improvement between the first two blocks is larger than the one of last two blocks. We conjecture that the first reattention is more accurate at measuring the similarity of word pairs by using the  original encoded word representation, while the latter reattention is distracted by highly nonlinear word representations. In addition, we notice that the attention deficiency has also been moderated: the divergence betwen normalized $E^t$ and ${E^t}^*$ is reduced.

\subsection{Prediction Analysis}
\begin{figure}
\begin{center}
\includegraphics[width=3.3in]{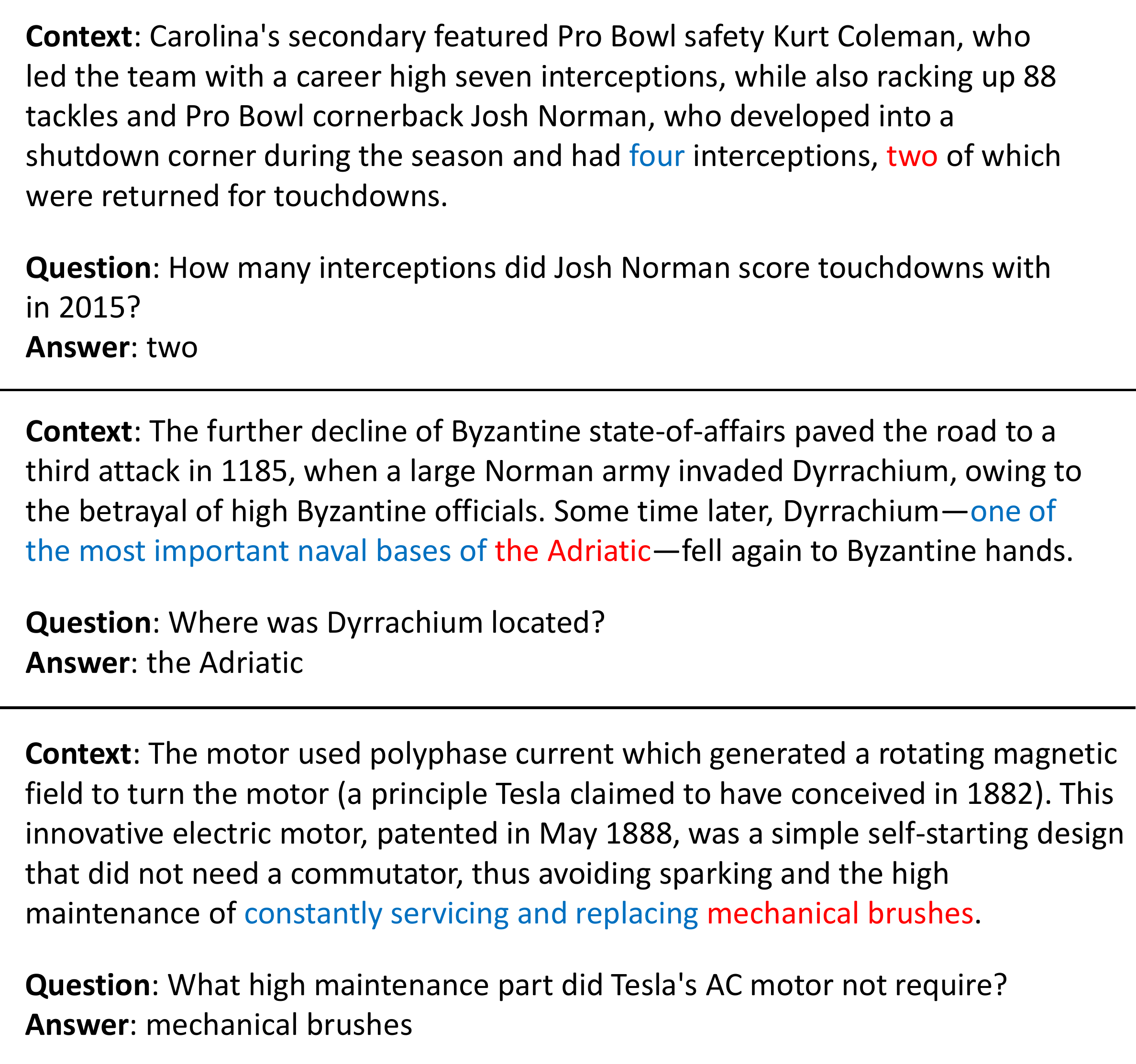}
\end{center}
\caption{Predictions with DCRL (red) and with SCST (blue) on SQuAD dev set.}
\label{fig3}
\end{figure}

Figure \ref{fig3} compares predictions made either with dynamic-critical reinforcement learning or with self-critical sequence training. 
We first find that both approaches are able to obtain answers that match the query-sensitive category. For example, the first example shows that both \emph{four} and \emph{two} are retrieved when the questions asks for \emph{how many}. 
Nevertheless, we observe that DCRL constantly makes more accurate prediction on answer spans, especially when SCST already points a rough boundary. 
In the second example, SCST takes the whole phrase after \emph{Dyrrachium} as its location. The third example shows a similar phenomenon, where the SCST retrieves the phrase \emph{constantly servicing and
replacing mechanical brushes} as its answer.
We demonstrates that this is because SCST encounters the convergence suppression problem, which impedes the prediction of ground truth answer boundaries.
DCRL, however, successfully avoids such problem and thus finds the exactly correct entity.

\section{Conclusion}
We propose the Reinforced Mnemonic Reader, an enhanced attention reader with two main contributions.
First, a reattention mechanism is introduced to alleviate the problems of attention redundancy and deficiency in multi-round alignment architectures.
Second, a dynamic-critical reinforcement learning approach is presented to address the convergence suppression problem existed in traditional reinforcement learning methods.
Our model achieves the state-of-the-art results on the SQuAD dataset, outperforming several strong competing systems. 
Besides, our model outperforms existing approaches by more than 6\% on two adversarial SQuAD datasets.
We believe that both reattention and DCRL are general approaches, and can be applied to other NLP task such as natural language inference. 
Our future work is to study the compatibility of our proposed methods.

\section*{Acknowledgments}
This research work is supported by National Basic Research Program of China under Grant No. 2014CB340303. 
In addition, we thank Pranav Rajpurkar for help in SQuAD submissions.

\bibliographystyle{named}
\bibliography{reference}

\end{document}